\documentclass[10pt,twocolumn,letterpaper]{article}

\usepackage{cvpr}              

\usepackage{microtype}

\renewcommand{\paragraph}[1]{\vspace{.5em}\noindent\textbf{#1.}}

\usepackage{xspace}
\newcommand{\paper}{\textbf{PlayClass}\xspace}

\usepackage{booktabs}
\usepackage{makecell}

\usepackage{soul}
\setuldepth{foobar}

\graphicspath{ {assets/} }
\usepackage{graphicx}
\usepackage{subcaption}

\definecolor{cvprblue}{rgb}{0.21,0.49,0.74}
\usepackage[pagebackref,breaklinks,colorlinks,allcolors=cvprblue]{hyperref}

\def\paperID{43} 
\def\confName{CVPR}
\def\confYear{2026}

\title{PlayClass: Automated Play Behaviour Classification in Poultry}

\author{Prince Ravi Leow$^{1*}$ \quad Neil Scheidwasser$^{1,3*}$ \quad Rebecca Oscarsson$^{2}$ \quad \\
Per Jensen$^{2\dagger}$ \quad Samir Bhatt$^{1,3\dagger}$ \quad David Alejandro Duch\^{e}ne$^{1\dagger}$\\[6pt]
$^{1}$Section for Health Data Science \& AI, University of Copenhagen\\
$^{2}$AVIAN Behaviour Genomics and Physiology Group, Link\"{o}ping University\\
$^{3}$Department of Infectious Disease Epidemiology, Imperial College London}

\begin{document}

\maketitle
\footnotetext{$^*$Equal contribution. $^\dagger$Supervising author.}

\begin{abstract}
    Automated monitoring of animal welfare has largely targeted negative indicators, leaving positive welfare behaviours such as play underexplored.
    To address this gap, we present \paper, a pipeline for play-behaviour classification in poultry from top-down pen video.
    The pipeline leverages long-duration tracking with SAM\,3 via YOLO-guided chunk boundaries to minimise identity errors in point-based prompting,
    and frozen embeddings from image and video foundation models for play action classification.
    Although handcrafted motion features from tracked masks alone achieved competitive accuracy, V-JEPA 2.1 consistently outperformed all other backbones across model scales,
    reaching 77.0 macro-averaged F$_1$ when combined with handcrafted features.
    Despite this result, the dataset remains challenging due to play sub-types sharing similar kinematic profiles with non-play and inter-bird occlusion.
    Overall, our work provides encouraging evidence towards automated frameworks for play behaviour classification in poultry.
\end{abstract}
\section{Introduction}
\label{sec:intro}
 
Precise measurement of animal behaviour is a fundamental problem in many areas in biology, including understanding neural mechanisms~\cite{anderson2014} and monitoring welfare in agricultural~\cite{dawkins2004} and conservation~\cite{sekar2020} settings.
Recent advances in computer vision have played a central role in this endeavour. Transfer learning approaches have greatly advanced the automation of animal pose estimation and tracking~\cite{mathis2020}. Vision foundation models~\cite{carion2025, simeoni2025, zhao2024, assran2025} pretrained on large, diverse datasets extend these capabilities with general-purpose representations that can be adapted effectively to specialised domains with limited labelled data. 
In agriculture, poultry constitutes an important case study. From a public health perspective, poultry farming is a cornerstone of food security and a major target of zoonosis surveillance~\cite{oecd2023}. From a technical perspective, poultry settings constitute a challenging vision domain, as they typically feature visually similar individuals in dense groups, which may lead to occlusions and identity confusion. 
Within poultry behaviour analysis, health and welfare assessment has been a dominant focus~\cite{dawkins2023}. A substantial body of work has leveraged optical flow methods to measure flock-level patterns of activity, impaired gait, or infection in commercial settings~\cite{dawkins2012, colles2016, donnelly2026}. Conversely, methods leveraging deep learning-based tracking are emerging for individual-level recognition of health-related markers (e.g., posture, feeding, sneezing) in experimental settings~\cite{cardoen2025, joo2022}.
However, these efforts have largely targeted \emph{negative} health and welfare indicators such as pain or disease~\cite{yeates2008}, leaving positive behaviours such as play~\cite{burghardt2005, held2011, smaldino2019} relatively underexplored. Play is a ubiquitous behaviour among animals, in spite of being energetically costly and physically risky~\cite{graham2010}. Despite its biological and welfare significance, automatic play
assessment is hindered by data scarcity and by definitional
ambiguity, as play often shares motor patterns with other behaviours and can transition rapidly between states~\cite{burghardt2005}.

\begin{figure*}[htbp]
  \centering
  \includegraphics[width=0.85\textwidth]{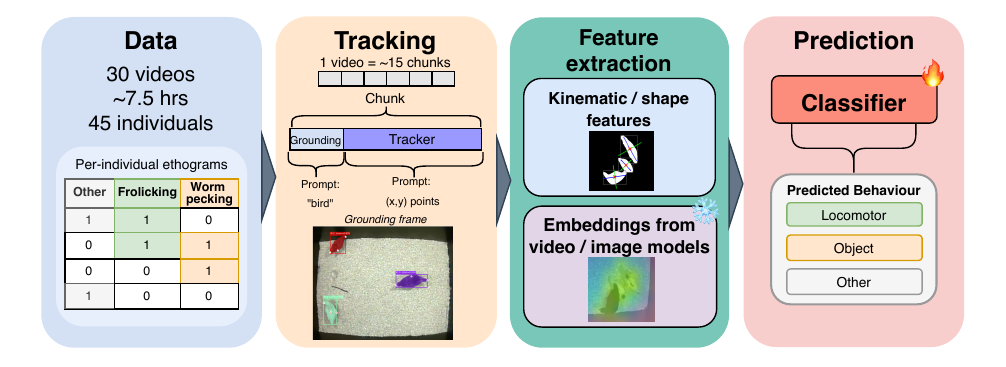}
  \caption{Overview of the \paper pipeline for play action classification in poultry, combining handcrafted motion features and frozen foundation-model embeddings extracted from 1.8 million SAM 3-tracked masks across 30 top-down video recordings.}
  \label{fig:pipeline}
\end{figure*}

In this work, we present \textbf{PlayClass}$^\ddagger$\footnotetext{$^\ddagger$\url{https://github.com/sbhattlab/PlayClassCV4Animals}} (Fig.~\ref{fig:pipeline}), a pipeline for individual-level play-behaviour classification in poultry from top-down video, and evaluate it on a dataset of 30 recordings covering 45 tracked individuals in a severely imbalanced regime.
Our contributions are:
\begin{enumerate}
    \item A tracking strategy that extends SAM 3~\cite{carion2025} to 15-minute recordings via YOLO-guided chunk-boundary selection and cross-chunk identity matching, maintaining individual identity through frequent occlusions.
    \item A benchmark of image (DINOv3~\cite{simeoni2025}) and video (V-JEPA 2/2.1~\cite{assran2025, murlabadia2026}, VideoPrism~\cite{zhao2024}) foundation models against a competitive baseline of handcrafted motion descriptors, where V-JEPA 2.1 emerged as the strongest backbone.
    \item An error analysis showing that misclassification is driven by dominant kinematic profiles within behaviour groups as well as occlusion scenarios.
\end{enumerate}
 
\section{Related Work}
\label{sec:related}
Open-source pose estimation tools such as DeepLabCut~\cite{mathis2018, lauer2022} have become a common starting point for behavioural analysis. Although pretrained pose models can generalise across dozens of species without manual labels~\cite{ye2024}, many scenarios still require \textit{de novo} keypoint labelling. As a label-light alternative, DINOv2~\cite{oquab2024} image embeddings from tracking masks have supported action classification in sheep~\cite{tibrewal2025}, while domain-adaptive pretraining of the V-JEPA~\cite{bardes2024} video foundation model has shown strong results for primate behaviour classification~\cite{mueller2025}.
V-JEPA 2.1~\cite{murlabadia2026} extends the latter with a dense predictive objective that grounds representations in fine-grained spatiotemporal structure, but has not yet been evaluated for animal behaviour.
For tracking, SAM\,3~\cite{carion2025} enables unprecedented promptable video segmentation, but remains difficult to apply to long clips ($>$1\,min) due to memory constraints and error accumulation~\cite{ding2025}.

Poultry-specific video data remains scarce, due in part to privacy and ownership constraints in agricultural settings. Recent efforts are beginning to address this gap, with datasets covering behaviour in general scenarios~\cite{joo2022, cardoen2025} and under infection~\cite{depuruAutomatingPoultryFarm2024, scheidwasser2025}. Play behaviour video has been collected~\cite{baxter2019, oscarsson2026} but without automated classification. To our knowledge, no prior work addresses automated play-behaviour classification in poultry.
\section{Methods}
\label{sec:method}

\subsection{Data}
From a controlled play study~\cite{oscarsson2026} on 45 Red
Junglefowl $\times$ White Leghorn chicks (housed in five cages of
nine), we used the 30 top-down video recordings
($704{\times}576$, 25\,fps, 15\,min each) collected on days~28
and~29 of rearing. On each recording day, cages were split into three focal groups of three, and each group was filmed separately in a dedicated arena provisioned with a fake worm and live mealworms to elicit object play.
The groups of three differed across days.
A trained observer scored play events per bird using 1/0 sampling in 5\,s windows according to an ethogram designed across two natural granularity levels: three play-behaviour categories (locomotor play, object play, social play) and 14 fine-grained sub-behaviours, detailed in Table~\ref{tab:ethogram}. We classify at the category level because the fine-grained scheme has severely uneven support (0 to 12{,}585 windows), with several sub-types too sparse to evaluate reliably under LOCO. Social play was excluded due to insufficient samples (201 windows, 1.4\%). The resulting dataset comprises 14{,}515 windows across 45 individuals, with severe class imbalance: 86.7\% other (non-play), 9.3\% object, 4.0\% locomotor.

\begin{table*}[t]
  \centering
  \scriptsize
  \caption{Ethogram of scored poultry play behaviours. $N$: number of labelled windows. Number of no-play windows: 12,585.}
  \label{tab:ethogram}
  \setlength{\tabcolsep}{2pt}
  \begin{tabular}{p{0.18\textwidth}p{0.76\textwidth}r}
      \toprule
      \textbf{Behaviour} & \textbf{Definition} & \textbf{$N$} \\
      \midrule
      \emph{Locomotor play} & & \textit{585} \\
      Frolicking           & Spontaneous and rapid running and/or jumping while wings either flapping or raised, often including rapid direction changes. & 531 \\
      Wing flapping        & Rapid vertical movement with both wings while stationary or walking up to 2 steps. \newline Excludes wing flaps performed by a bird to balance itself or correct its feathers. & 25 \\
      Running              & Spontaneous forward movement at $\geq 2\times$ normal walking pace, often including rapid direction changes. \newline Can start either while walking or being stationary. Wing flapping does not occur simultaneously. & 22 \\
      Spinning             & Circling movement around the bird's own axis with $\geq 2\times$ the normal walking pace. Wing flapping does not occur simultaneously. & 0 \\
      Spinning while wing flapping & As above, but wing flapping occurs simultaneously. & 7 \\
      \midrule
      \emph{Object play} & & \textit{1345} \\
      Worm pecking         & Bird pecks at fake worm or mealworm on the ground. The worm may be lifted off the ground and/or shaken but not carried. & 835  \\
      Object running       & A bird picks up an object (normally the fake worm) in its beak and starts either walking or running. \newline Rapid changes of pace or direction might occur. Other birds may begin to chase the bird carrying the object. & 279  \\
      Worm running         & As above, with a live mealworm. & 99 \\
      Object/worm chasing  & A bird chases after an individual performing object- or worm-running and may try to obtain the carried object or worm. & 131 \\
      Object/worm exchange & An object or worm is obtained by one bird from another bird's beak. \newline The object can be obtained either by a chasing bird from a running bird, or from a stationary bird with an object in its beak. & 1 \\
      \midrule
      \emph{Social play (excluded)} & Simulation of fighting behaviour with no obvious aggression. & \textit{201} \\ 
      Sparring jumping, no contact &  Two birds jumping while standing close facing one another. May be performed by one bird while the other is passive or lightly avoidant. Wings flapping, extended, or kept to the sides. May include light kicking. No physical contact involved. & 11 \\
      Sparring jumping, with contact & As above, but involving physical contact. & 17 \\
      Sparring stand-off, no contact & Birds face each other briefly, stepping close and backing off from one another. Necks and feathers around the neck are often raised. \newline Pecking towards neck, head, or beak of the receiving bird. Wings flapping, extended, or kept to the sides. No physical contact involved. & 161 \\
      Sparring stand-off, with contact & As above, but involving physical contact. & 12 \\
      \bottomrule
  \end{tabular}
\end{table*}

\subsection{Tracking}
\label{sec:tracking}
For each video, we performed multi-animal tracking with SAM\,3~\cite{carion2025} to produce per-frame segmentation masks for each bird. Because GPU memory limited tracking each 15-min recording end-to-end, we processed videos in chunks of $~60$ s. Tracking was initialised with the text prompt \emph{bird}, and subsequent chunks were prompted using points extracted from the previous chunk's final masks. To mitigate error accumulation over long sequences~\cite{ding2025}, we implemented an \emph{adaptive chunking} strategy, where a YOLO26x~\cite{jocher2026} detector + BoT-SORT~\cite{aharon2022} tracker were used to place chunk boundaries at frames maximising inter-bird separation, thereby facilitating identity transfer across chunks. An additional \textit{adaptive grounding} strategy was implemented to improve initialisation: instead of simply starting at the first frame, SAM\,3 was run with the text prompt over the first 5\,s (125 frames) to identify a grounding frame with well-separated and high-confidence detections.

We evaluated the tracker on sparse human-verified keyframes from five videos ($\sim$90 frames each, 462 in total), selected to cover challenging scenes (crowding, occlusion). Table~\ref{tab:tracker_eval} reports HOTA~\cite{luitenHOTAHigherOrder2021} and IDF$_1$~\cite{ristaniPerformanceMeasuresData2016}, showing that SAM\,3 outperformed common tracking baselines. Both adaptive chunking and adaptive grounding yielded measurable gains over their default alternatives. However, even with these improvements, manual post-processing was required to resolve cross-chunk ID switches, correct tracking anomalies, and map tracker IDs to the experimental protocol IDs. The final tracked dataset comprises over 1.8\,million masks with bounding boxes.

  \begin{table}[t]
      \centering
      \small
      \setlength{\tabcolsep}{6pt}
      \caption{Tracking evaluation and ablation (mean HOTA and IDF$_1$ $\pm$ SD aggregated over sparse human-verified keyframes from five videos). The best score is shown in \textbf{bold}.}
      \label{tab:tracker_eval}
      \begin{tabular}{lcc}
          \toprule
          \textbf{Tracking method} & \textbf{HOTA} & \textbf{IDF$_1$} \\
          \midrule
          \multicolumn{3}{l}{\emph{Baselines:}} \\
          YOLO26x + BoT-SORT & 0.065 \scriptsize{$\pm$0.016} & 0.059 \scriptsize{$\pm$0.022} \\
          Grounded-SAM-2~\cite{renGroundedSAMAssembling2024} & 0.282 \scriptsize{$\pm$0.196} & 0.308 \scriptsize{$\pm$0.263} \\
          \quad + adaptive grounding & 0.349 \scriptsize{$\pm$0.064} & 0.482 \scriptsize{$\pm$0.099} \\
          \midrule
          \multicolumn{3}{l}{\emph{Proposed method:}} \\
          \makecell[l]{SAM 3 + adaptive grounding \\ \quad \& adaptive chunking}
            & \textbf{0.563} \scriptsize{$\pm$0.057} & \textbf{0.700} \scriptsize{$\pm$0.116} \\
          \midrule
          \multicolumn{3}{l}{\emph{Ablations on proposed method:}} \\
          \quad $-$ adaptive grounding & $-$0.275 \scriptsize{$\pm$0.060} & $-$0.335 \scriptsize{$\pm$0.120} \\
          \quad $-$ adaptive chunking  & $-$0.013 \scriptsize{$\pm$0.130} & $-$0.030 \scriptsize{$\pm$0.197} \\
          \bottomrule
      \end{tabular}
  \end{table}

\subsection{Feature Extraction}
\label{sec:features}

From the post-processed video tracks obtained via SAM\,3, we extracted two complementary feature streams per bird per observation window.

\paragraph{Handcrafted mask features}
Nineteen per-frame features capturing spatial shape properties (e.g., mask area, solidity, circularity), temporal dynamics (e.g., velocity, acceleration, turning angle), and pairwise social context (distances to other tracked birds) were computed from the tracked masks. 
Each feature was summarised per ethogram window using nine summary statistics (e.g., moments, percentiles), yielding a 171-dimensional vector per window.

\paragraph{Visual embeddings}
We evaluated frozen image and video foundation models as embedding backbones.
For image models (DINOv3~\cite{simeoni2025}), per-frame CLS-token embeddings were extracted from tight bounding-box crops resized to the model's native resolution.
For video models (V-JEPA 2~\cite{assran2025}, 2.1~\cite{murlabadia2026}, VideoPrism~\cite{zhao2024}), non-overlapping clips of $K_\text{in}$ frames were processed and spatially mean-pooled per timestep,
then concatenated across clips.
Both approaches yield a variable-length sequence of size $F_w \times D$ per window, where $F_w$ is the number of temporal tokens and $D$ the embedding dimension.
We compare ViT-B and ViT-L variants across backbone families (see \hyperref[sec:experiments]{Experiments}).

\subsection{Classification}
\label{sec:classifier}

Our problem consists of a 3-class classification task over \emph{locomotor play}, \emph{object play}, and \emph{other} (no play).
Play events were much rarer (object 9.3\%, locomotor 4.0\%) than the majority class (86.7\%).

\paragraph{Classifier architectures}
Two simple classification architectures were implemented: a multi-layer perceptron (MLP) probe operating on mean-pooled embeddings or windowed feature statistics,
and a 1D convolutional network (1D-CNN) to evaluate the value of keeping temporal signal.
For the latter, variable-length embedding sequences were first compressed into $K$ fixed-length segments via adaptive average pooling, where $K$ was set to match each backbone's temporal granularity (see \hyperref[sec:experiments]{Experiments}) or tuned in downstream experiments (DINOv3).
Subsequently, the segments were projected through a GELU~\cite{hendrycks2023} 
bottleneck, a single 1D convolution layer, and gated attention pooling~\cite{ilse2018} before a linear classification head.

\paragraph{Training}
To obtain a more objective estimate of generalisation performance, we used Leave-One-Cage-Out (LOCO) cross-validation with five folds (one per cage).
To prevent environment leakage, in each fold, one cage was held out as the test set, while the validation set comprised videos from the next cage in circular order, with the remaining three cages used for training.
All models were trained for five epochs using AdamW~\cite{loshchilov2019} 
with weighted cross-entropy loss (inverse-square-root class weights) and label smoothing ($\alpha{=}0.1$).
The checkpoint with the best validation loss was selected for testing.
We report the macro-averaged F$_1$ score, computed from the aggregated confusion matrices across folds.

\section{Experiments}
\label{sec:experiments}

To understand the drivers of play-classification performance in our data-scarce setting, we conducted ablation studies over backbone models, input representations, and training hyperparameters.
In hybrid configurations combining handcrafted features with frozen embeddings, windowed feature statistics were concatenated with the 1D-CNN's representation before the classification head.
Importantly, fold-level standard deviations were uniformly high across all experiments ($\pm$4--6\%), reflecting variation in play frequency across cages.
Due to strong class imbalance in the dataset, we used the macro-averaged F$_1$ score as the primary summary metric (Table~\ref{tab:ablation}) and additionally report class-level precision, recall, and F$_1$ for the best model (Table~\ref{tab:per-class}).


Table~\ref{tab:ablation} compares frozen foundation model backbones, handcrafted features, and their combination.
Since ViT-L is the only architecture shared by all four backbone families, we use it as the primary comparison point.

Handcrafted features alone achieved 73.4 macro-averaged F$_1$ with a simple MLP, establishing a strong baseline without any learned visual representations.
All frozen embedding backbones with the 1D-CNN surpassed this baseline at ViT-L scale, confirming that learned visual features capture information beyond linear kinematic and shape descriptors.

Among backbones, V-JEPA 2.1 consistently outperformed all alternatives at both ViT-B and ViT-L scales.
The advantage was driven by higher object play recall in V-JEPA 2.1 (59.0\% vs 53.4\% for DINOv3), while locomotor play recall was comparable across backbones (recall ${\sim}$75\%).

Despite strong scaling on standard benchmarks~\cite{simeoni2025, zhao2024, murlabadia2026}, increasing model size yielded minimal gains in our setting, with V-JEPA 2.1 ViT-B surpassing all other backbones at any scale.
Scaling DINOv3 to ViT-H did not improve over ViT-L (macro-averaged F$_1$: 73.4 vs 74.0), suggesting that representations saturate early for this domain-specific task.
Combining handcrafted features with V-JEPA 2.1 ViT-L yielded the best result of \textbf{77.0} macro-averaged F$_1$.

Rather than showing a large performance gap over handcrafted descriptors, these results show that handcrafted mask features constitute a strong and interpretable baseline for poultry play recognition. This indicates that while much of the task-relevant signal appears to be captured by motion, shape and proximity statistics, frozen video representations provide complementary information.

\begin{table}[t]
  \centering
  \small
  \caption{Backbone evaluation and training ablation (macro-averaged F$_1$ $\pm$ SD across LOCO folds). All embedding rows use the 1D-CNN classifier. The best score is shown in \textbf{bold}.}
  \label{tab:ablation}
  \begin{tabular}{lccc}
      \toprule
      \textbf{Input} & \textbf{$K$} & \textbf{ViT-B} & \textbf{ViT-L} \\
      \midrule
      Features (MLP)         & --- & \multicolumn{2}{c}{73.4 \scriptsize{$\pm$4.7}} \\
      \midrule
      DINOv3                 & 24 & 70.7 \scriptsize{$\pm$5.5} & 74.0 \scriptsize{$\pm$6.7} \\
      VideoPrism             & 8/16 & 73.8 \scriptsize{$\pm$4.4} & 74.1 \scriptsize{$\pm$4.9} \\
      V-JEPA 2               & 32 & ---   & 74.3 \scriptsize{$\pm$4.5} \\
      V-JEPA 2.1             & 32 & 75.8 \scriptsize{$\pm$5.0} &  76.3 \scriptsize{$\pm$5.4} \\
      \midrule
      Feat.\ + V-JEPA 2.1   & 32 & 76.5 \scriptsize{$\pm$4.1} & \textbf{77.0} \scriptsize{$\pm$5.5} \\
      \midrule
      \multicolumn{4}{l}{\emph{Ablations on best model:}} \\
      \quad 1 epoch                    & 32 & --- & -4.8 \scriptsize{$\pm$5.4} \\
      \quad $-$ class weights          & 32 & --- & -1.5 \scriptsize{$\pm$4.5} \\
      \quad $-$ label smoothing        & 32 & --- & -1.5 \scriptsize{$\pm$4.0} \\
      \quad $-$ 1D-CNN (MLP)           & --- & --- & -0.9 \scriptsize{$\pm$4.7} \\
      \bottomrule
  \end{tabular}
\end{table}

The lower portion of Table~\ref{tab:ablation} ablates the training recipe. Beyond training epochs, components for class imbalance mitigation (inverse square root class weighting, label smoothing) were most impactful. Surprisingly, replacing the 1D-CNN with an MLP did not significantly degrade performance, suggesting that mean-pooled V-JEPA 2.1 embeddings already capture most temporal information for short events. Lastly, coarsening the temporal resolution from $K = 32$ to $K = 16$ segments per window degraded performance by 1.7, but finer resolution ($K = 48$) offered no further gain.

\begin{table}[t]
  \centering
  \small
  \caption{Row-normalised confusion matrix (\%) with marginal
  precision and per-class F$_1$ for the best model.}
  \label{tab:per-class}
  \begin{tabular}{@{}lccc|cr@{}}
      \toprule
      & \multicolumn{3}{c|}{\textbf{Predicted}} & & \\
      \textbf{True} & \emph{Other} & Object & Loc. & \textbf{F$_1$} & \textbf{Support} \\
      \midrule
      \emph{Other}  & 93.5 &  4.8 &  1.7 & 94.8 & 12{,}585 \\
      Object        & 31.2 & 66.1 &  2.7 & 61.9 &  1{,}345 \\
      Loc.          &  8.5 &  6.7 & 84.8 & 74.4 &    585 \\
      \midrule
      \textbf{Prec.}& 96.2 & 58.2 & 66.2 & & \\
      \bottomrule
  \end{tabular}
\end{table}

\subsection{Error Analysis}
\label{sec:analysis}

\begin{figure}[t]
    \centering
    \includegraphics[width=\linewidth]{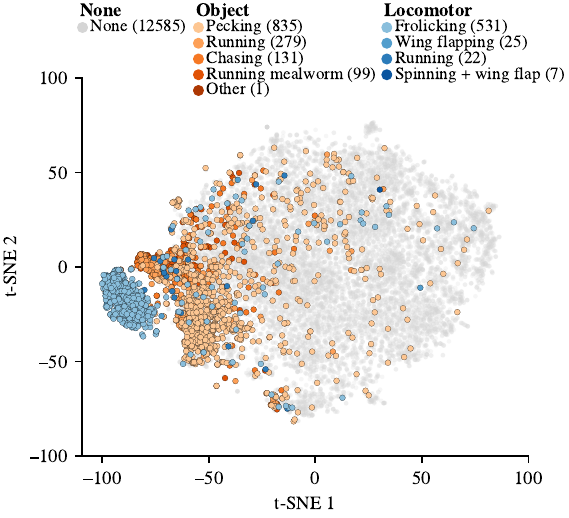}
    \caption{t-SNE projection of feat. + V-JEPA 2.1 ViT-L embeddings coloured by fine-grained behaviour.}
    \label{fig:tsne}
\end{figure}

\paragraph{Sub-behaviour ambiguity} Despite class-imbalance mitigation, all models generally retained a bias towards the majority class.
For the best model, object play had
the lowest recall (66.1\%) and precision (58.2\%) of any class,
with most misclassifications going to the majority non-play class  (Table~\ref{tab:per-class}).
This bias was amplified by crowding (Spearman $\rho = 0.17$ between
bird-bird distance and object-play accuracy, $p < 10^{-9}$). We further probe this quantitatively in the embedding geometry, projected using t-SNE in Fig.~\ref{fig:tsne}. Although distinct clusters for each major play behaviour exist, sub-clusters driven by shared kinematic profiles emerge across categories. k-nearest neighbour probing (k-NN; k = 1) of the model's embeddings shows that 57\% of worm pecking windows lie closest to a non-play window, reflecting their diffuse t-SNE placement. Active object play sub-types such as worm-running (90\% recall) and worm-chasing (60\%) form more distinct clusters, though with partial overlap with locomotor behaviours (wing flapping, running), reflecting shared linear locomotion profiles across play categories (36\% of Running windows have an object-play nearest neighbour). Conversely, frolicking forms a well-separated cluster (88\% recall; 71\% self-neighbours), potentially indicative of its distinctive kinematic signature of rapid, multi-directional movement~\cite{baxter2019}. Overall, these patterns echo the definitional ambiguity of play behaviour~\cite{burghardt2005}: similar motor patterns can be functional rather than playful, with rapid transitions between play and non-play states.

\begin{figure}[t]
    \centering
    \includegraphics[width=0.8\linewidth]{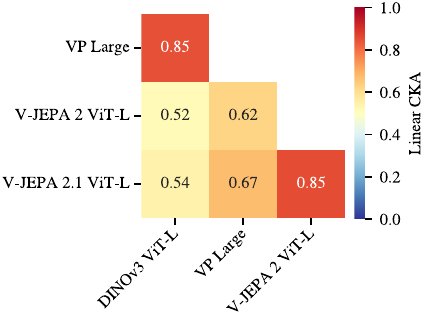}
    \caption{Linear CKA similarity between backbone representations (mean-pooled per window). VP: VideoPrism~\cite{zhao2024}.}
    \label{fig:cka}
\end{figure}

\paragraph{Representation similarity}
Centred kernel alignment analysis~\cite{kornblith2019} (CKA; Fig.~\ref{fig:cka}) was used to compare the per-window representations produced by the four ViT-L foundation models. The obtained similarity matrix generally grouped backbones by training paradigm, with DINOv3/VideoPrism and V-JEPA 2/V-JEPA 2.1 both at CKA $\approx$ 0.85 and cross-family CKA ranging from 0.5 to 0.7. The convergence of DINOv3 and VideoPrism may reflect their shared reliance on semantic teacher targets, potentially inducing biases toward appearance over motion. Conversely, the motion-predictive structure of V-JEPA models may better capture play kinematics. Beyond model similarity, the weaker classification performance of VideoPrism could be due to its shorter temporal context (8--16 frames vs. 64 for V-JEPA), limiting its capacity to capture the sustained motion patterns characteristic of play.

\section{Conclusion}
\label{sec:conclusion}
We present \paper, a pipeline for play-behaviour classification in poultry, including SAM 3-based tracking and play action classification with image and video foundation models.
Among frozen foundation models, V-JEPA 2.1 consistently outperformed image-based and other video backbones, achieving 77.0 macro-averaged F$_1$ when combined with handcrafted motion features. 

Models discriminated well between behaviours with distinct kinematic profiles, but still struggled in occlusion scenarios and with ambiguous sub-types where play and non-play share similar motor patterns. Surprisingly, handcrafted features alone reached 73.4 macro-F$_1$, only 3.6 points below the best hybrid (77.0). Thus, play recognition may largely be captured by kinematic cues, with frozen V-JEPA 2.1 embeddings contributing complementary signal. Beyond model failure modes, our pipeline currently relies on manual identity correction in post-processing for downstream classification, leaving a gap to
fully autonomous deployment.
Future work could thus leverage larger poultry behaviour datasets~\cite{cardoen2025} for domain-adaptive pretraining, incorporate cross-bird spatial context to better handle occlusion and social interactions, and improve cross-chunk identity matching to reduce reliance on manual correction. 
Overall, we believe that automating play-behaviour detection can enable welfare assessment at scales previously impossible, contributing to evidence-based improvements in livestock management.
\clearpage
{
    \small
    \bibliographystyle{ieeenat_fullname}
    \bibliography{main}
}


\end{document}